# Enhancing Mathematics Learning for Hard-of-Hearing Students Through Real-Time Palestinian Sign Language Recognition: A New Dataset

Fidaa khandaqji[1], Huthaifa I. Ashqar[2,3], Abdelrahem Atawnih[1]

[1] Department of Natural, Engineering and Technology Sciences, Arab American University, Jenin P.O. Box 240, Palestine
`f.khandaqji@student.aaup.edu`
`abdelrahem.atawnih@aaup.edu`

[2] Department of AI and Data Science, Arab American University, Jenin P.O. Box 240, Palestine
`Huthaifa.ashqar@aaup.edu`

[3] AI Program, Columbia University, New York, NY 10027, USA

**Abstract.** The study aims to enhance mathematics education accessibility for hard-of-hearing students by developing an accurate Palestinian sign language PSL recognition system using advanced artificial intelligence techniques. Due to the scarcity of digital resources for PSL, a custom dataset comprising 41 mathematical gesture classes was created, and recorded by PSL experts to ensure linguistic accuracy and domain specificity. To leverage state-of-the-art-computer vision techniques, a Vision Transformer ViTModel was fine-tuned for gesture classification. The model achieved an accuracy of 97.59%, demonstrating its effectiveness in recognizing mathematical signs with high precision and reliability. This study highlights the role of deep learning in developing intelligent educational tools that bridge the learning gap for hard-of-hearing students by providing AI-driven interactive solutions to enhance mathematical comprehension. This work represents a significant step toward innovative and inclusive frosting digital integration in specialized learning environments. The dataset is hosted on Hugging Face at https://huggingface.co/datasets/fidaakh/STEM_data.

## 1 Introduction

Sign language is a fundamental communication tool for the deaf and hard–of–hearing, processing its grammatical structure and complexities. According to the World Health Organization (WHO), over 430 million people worldwide experience some degree of hearing loss, a number expected to reach one billion by 2050, with at least 700 million requiring support [1].

Across the globe, more than 300 different sign languages are in use, each exhibiting unique syntactic and semantic properties that differentiate them from both spoken languages and one another. Notable examples include American sign language ASL, British sign language BSL, Arabic sign language ArSL, and Chinese sign language CSL. Despite significant technological advancements, real-time sign language recognition remains a formidable challenge due to linguistic diversity, regional dialects, and the complexity of manual and non-manual components of sign language. Traditional approaches to sign language translation have heavily relied on human interpreters, creating bottlenecks in accessibility. However, the latest advancements in automatic recognition techniques have revolutionized this field replacing expert assistance with AI-powered models capable of translating sign gestures into text and speech in real-time, ensuring greater accessibility for users [2,3,4].

Despite recent advancements, many existing solutions still face significant limitations, particularly in supporting under-resourced and often overlooked sign languages such as Palestinian sign language. Unlike widely studied sign languages like ASL, and BSL, PSL has received minimal academic attention, leaving a significant gap in research and technological development. PSL plays a crucial role in communication and education for the 76,480 deaf and hard-of-hearing individuals in Palestine, yet only 1,700 students are enrolled in public schools, highlighting severe educational disparities [5].

Recent advancements in sign language recognition have primarily centered on static gesture classification. However, sign language inheritance is dynamic, requiring models capable of processing movements. Early studies focused on spatial analysis using CNN-based architecture, but these approaches were limited in capturing temporal dependencies. The introduction of RNNs, particularly LSTMs, enhanced the ability to process sequential data, improving recognition accuracy in continuous sign language [6]. Notably, the YOLO5 framework has demonstrated high efficiency in real-time Arabic sign language recognition [7], while region-based CNN (R-CNN) has shown remarkable accuracy in classifying Arabic sign language letters [8]. Further advancements have been made using CNN and LSTM networks, which have enhanced recognition accuracy for Egyptian sign language ESL) despite challenging environmental conditions [9]. Other approaches have incorporated depth data and sensor-based recognition, further refining gesture accuracy and real-time processing, and offering higher accuracy in recognizing manual movements through time and spatial data integration [10,11,12].

Similarly, a study on Indian sign language leveraged LSTM networks to improve dynamic sign language, demonstrating enhanced flexibility and precision [13]. Computer vision techniques have been explored for Indonesian sign language, revealing variations in performance when transitioning from image-based to video-based processing models [14]. Similarly, multimodal approaches combining skeletal tracking and human Key Point extraction have improved recognition across languages like Korean sign language KSL and Turkish sign language TSL [15,16,17]. Moreover, Japanese and Indian sign language recognition systems have benefited from motion-sensor technology, improving real-time performance [19,18]. Additionally, machine learning–based classification of Indian sign language gestures using MATLAB has enhanced sign language detection accuracy for single- and double-hand movements [20].

One of the most promising developments in SLR is the application of ViTs. These models excel in capturing spatial and temporal dependencies, offering superior performance compared to traditional CNN-based architecture. A ViT-based model for Indian sign language achieved 99.29% accuracy with minimal training epochs [21]. Similarly, ViTs have outperformed CNNs in Arabic sign language recognition, proving more effective in gesture classification and feature extraction [22]. Additionally, self–supervised video transformers have been explored for Isolated sign language recognition, providing enhanced feature representations in gesture analysis [23].

Another major limitation in current SLR models is their heavy reliance on hand gesture recognition alone, while neglecting facial expressions and body movements, both of which are crucial for accurate PSL interpretation [24]. Additionally, while deep learning architecture has greatly improved classification accuracy, it often requires high computational power, making real-time deployment impractical for low-resource environments [25].

Despite advancements in sign language recognition, significant gaps remain, particularly for the Palestinian language. Most studies focus on well-documented sign languages such as ASL, and BSL. Neglecting PSL's unique linguistic classification, overlooking dynamic sign sentences, facial expression, and real-time interaction, limiting practical applications in education and assistive communication.

This study's key contributions by developing a specialized dataset for Palestinian sign language PSL, focusing on mathematical concepts to address the scarcity of linguistic resources it also highlights the role of artificial intelligence in enhancing education for the hard-of-hearing community.

## 2  Methodology

This study is to develop a robust sign language recognition system for PSL specifically targeting mathematical concepts to improve accessibility in education for hard-of-hearing students. The methodology consists of several key stages. Including data collection, model selection and training, evaluation, and system implementation.

### 2.1  Data Collection and Preprocessing

The dataset preparation for this study was influenced by PSL-specific grammatical patterns identified in prior linguistic research [26]. Given that PSL relies heavily on non-manual markers (e.g. facial expressions, head tilts, and spatial organization), the selection of gestures for the AI models was based on structural consistency with PSL, grammar rather than simply mapping signs to spoken language equivalents. The dataset of PSL was specifically curated to focus on mathematical concepts, including numbers, arithmetic operations, and geometric shapes. It consisted of 41 gesture videos, each representing a distinct class. These videos were recorded by 21 proficient volunteers in PSL. Ensuring diverse representation of gestures. The gestures were sourced and standardized based on the Palestinian sign language dictionary, originally published by the Palestinian Red Crescent Society in 1991 and updated in 2021. This dictionary served as a reliable reference for accurate and consistent gestures. Figure 1 below represents the sample images from the PSL dataset showing various sign language gestures used for mathematical concepts, including numbers and operations such as addition, subtraction, multiplication, and gesture shapes.

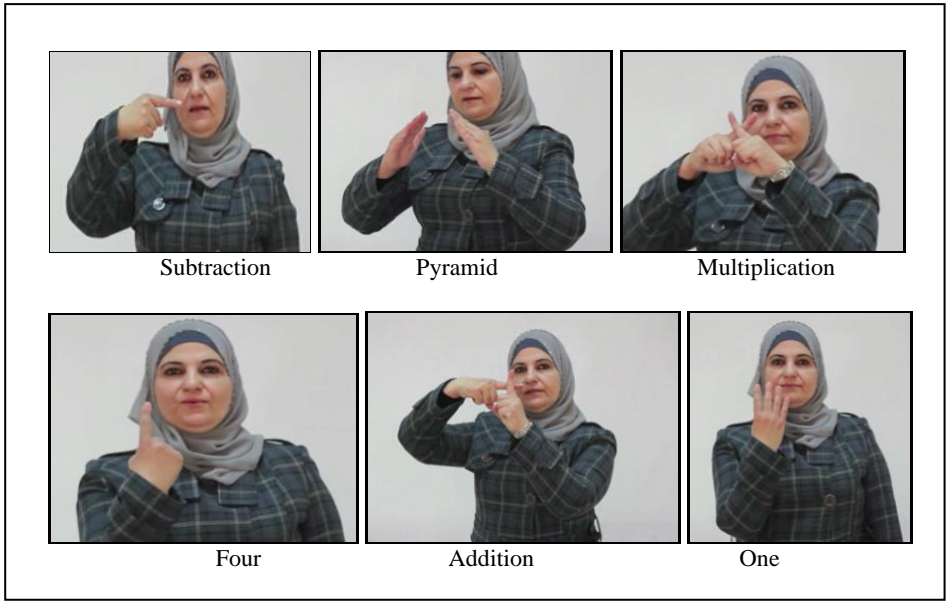

**Figure 1**: sample images from the PSL dataset.

The dataset was divided into three subsets:

1. Training data: 24 videos 2316 frames, 80% of the dataset were used for Training the vision transformer (ViT) model, providing a comprehensive set of examples for the model to learn from.
2. Validation data: Nine videos and 289 frames, 10% of the dataset, were used for validation during training to monitor the model's generalization performance and avoid overfitting.
3. Testing data: Eight videos, 290 frames. Ten percent of the dataset was reserved for final evaluation, ensuring an unbiased assessment of the model's classification capabilities.

In total, the dataset comprised 2896, ensuring adequate representation of all gesture classes while maintaining a balanced distribution across the subsets. To enhance model robustness and generalization, various data augmentation techniques were applied, including Random horizontal flipping to account for natural hand variations, rotation transformations to handle slight gesture angle variations, color jittering to improve adaptability to different lighting conditions, normalization, and resizing standardized for model training. All frames were extracted from the video and used in training, ensuring that the model captured the full sequence of movements for accurate sign recognition. Extract frames were resized to 224*224 pixels to match the input size required by the ViT model. Additionally, pixel values were normalized to the [0.1] range to maintain consistency and improve convergence during training.

Figure 2 illustrates the distribution of gesture classes in the training dataset. Each numeric label on the horizontal axis represents a specific sign. The vertical axis reflects the number of frames extracted from videos of each sign class. While the number of frames per class varies slightly, the overall distribution remains relatively balanced, which helps ensure fair and consistent training of the model.

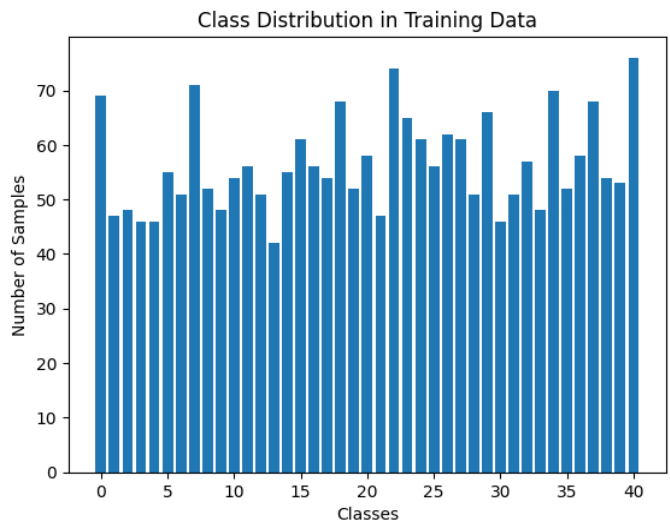

**Figure 2:** Class Distribution of Training Data.

## 2.2 Model Architecture and Experimental Setup

Figure 3 illustrates the architecture of the vision transformer ViT model used for gesture classification in this study. The input video is decomposed into image frames, which are then split into patches. These patches are linearly projected and enriched with positional embeddings before being passed through the transformer encoder. The self-attention mechanism enables the model to capture complex spatial dependencies across the patches. The output is passed through a multi-layer perception (MLP) head for final classification into one of 41 gesture classes. The model used is ViT- base-patch 16-224, fine-tuned on the custom PSL dataset with pre-trained weights from ImageNet. To ensure effective training and evaluation, we selected the ViT- base-patch 16-224 model due to its proven success in image classification tasks and its ability to capture long-range dependencies through the self-attention mechanism. This architecture offers a balanced trade-off between computational complexity and accuracy, making it a suitable choice for our mid-sized dataset focused on gesture recognition.

The implementation was carried out using PyTorch and the Hugging Face Transformers library, which provide a flexible, modular, and widely adopted framework for

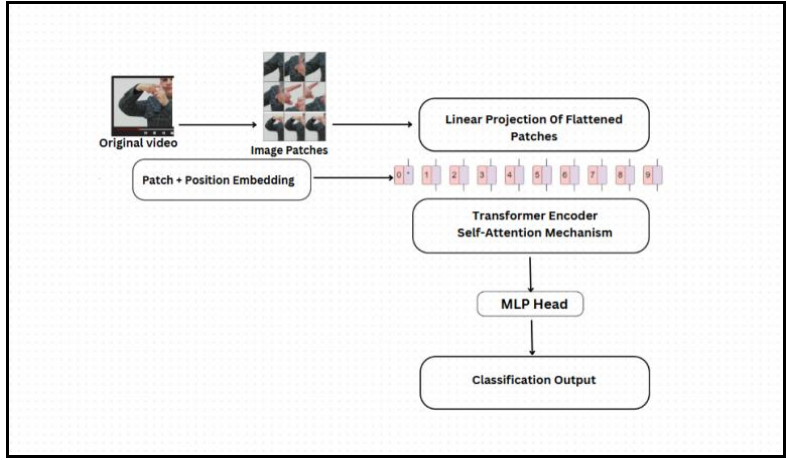

**Figure 3:** Vision Transformer Architecture for PSL Gesture Recognition.

transformer-based model development. These tools also facilitate seamless integration with pre-trained weights and offer utilities for efficient fine-tuning experimentation

All experiments were conducted on Google Colab Pro, utilizing an NVIDIA A100 GPU to accelerate training and support the computational demands of the ViT model. For video preprocessing and frame extraction, we employed OpenCV, given its robust performance, ease of use, and board compatibility with different video formats.

Table 1 summarizes the key hyperparameters and settings used in the ViT-base-patch16-224 on the PSL dataset. The model was fine-tuned using the PyTorch framework along with the hugging face transformers library. The optimizer AdamW was selected for its improved generalization over standard Adam, and a learning rate of 2 e-5 was found to be effective for stable fine-tuning. A batch size of 23 and 10 training epochs was chosen to balance training time and performance. The cross-entropy loss function was used as it is suitable for multi-class classification problems.

**Table 1.** Training Configuration

| Parameter | Value |
| --- | --- |
| Optimizer | AdamW |
| Learning Rate | 2e-5 |
| Batch Size | 32 |
| Epochs | 10 |
| Loss Function | Cross Entropy Loss |

**Table 1**: Training configuration hyperparameters used for fine-tuning the ViT-base- patch16-224 model on the PSL gesture dataset.

The following pseudocode Algorithm 1 outlines the step-by-step procedure used in the system, including video frame extraction and preprocessing. Patch embedding, and classification through a transformer encoder. This structured approach ensures efficient and accurate recognition of PSL gestures, enabling seamless interaction and learning experiences.

| Algorithm 1 Palestinian Sign Language Recognition Using ViT |
|---|
| Input: Video containing PSL gestures |
| Output: Recognized sign class |
| 1. Procedure load_video (video_path) |
| 2.    Frames ←extract_frames (video_path, frame_rate) |
| 3. Return frames |
| 4. End procedure |
| 5. Procedure preprocess_frame (frame) |
| 6.    frame←RESIZE (frame, (224,224)) |
| 7.    frame← Normalize (frame, mean= [0.5,0.5,0.5], std= [0.5,0.5,0.5]) |
| 8.    frame←TO_TENSOR (frame) |
| 9.    return frame |
| 10. end procedure |
| 11. procedure extract_patches (frame, patch_size) |
| 12.    patches←SPLIT (frame, patch_size) |
| 13.    Return patches |
| 14. end procedure |
| 15. procedure transformer_encoder (patches) |
| 16.    encoded_patches ←LINER_PROJECTION (patches) |
| 17.    Position_embeddings← ADD_POSITIONAL_ENCODING (encoded_patches) |
| 18.    Attention_output←SELF_ATTENTION (position_embeddings) |
| 19.    Return attention_output |
| 20. end procedure |
| 21. procedure ViT_classification (video) |
| 22.    frames← load_video(video) |
| 23.    For each frame in frames do |
| 24.       Processed_frame← preprocess_frame(frame) |
| 25.       patches←extract_patches (processed_frame, patch_size=16) |
| 26.       features← tracnsformer_encoder(patches) |
| 27.    End for |
| 28.    Classification_result←MLP (features) |
| 29.    Return classification_result |
| 30. end procedure |

## 2.3 Model Evaluation

To Evaluate the performance of the proposed ViT-based system for PSL recognition, we employed four key metrics: Accuracy, Precision, Recall, and F1_score. These metrics provide a comprehensive understanding of the model's effectiveness in classifying gestures accurately, especially when dealing with imbalanced datasets or overlapping gesture classes. Accuracy was used to measure the model's overall correctness by calculating the ratio of correctly predicted samples both positive and negative to the total number of samples. While accuracy provides a board overview, it can be misleading in cases of imbalanced data. Therefore, we incorporated precision to evaluate the model's ability to correctly classify positive cases without false positives and Recall to assess the model's sensitivity in detecting all true positive cases, finally, the F1-score, which is the harmonic mean of precision and Recall, was used to provide a balanced measure of the model's performance, especially in scenarios where both metrics are equally important.

$$Accuracy = \frac{TP+TN}{TP+TN+FP+FN} \quad (1)$$

$$Precision = \frac{TP}{TP+FP} \quad (2)$$

$$Recall = \frac{TP}{TP+TN} \quad (3)$$

$$F1\_score = 2 \cdot \frac{precision \cdot recall}{precision+recall} \quad (4)$$

Where:

TP: the number of correctly predicted positive instances.
TN: the number of correctly predicted positive instances.
FP: the number of incorrectly predicted positive instances.
FN: the number of incorrectly predicted negative instances.

$$l = -\frac{1}{N} \sum_{i=1}^{N} \sum_{C=1}^{N} yi, c \, \log(yi, c) \quad (5)$$

Where: N: number of samples, C: number of classes, yi, c: ground truth table.

## 3 Results and Discussion

The ViT model demonstrated exceptional learning capabilities during training on the PSL dataset, which consisted of 41 gesture classes derived from videos of mathematics concepts. The training and validation losses converged effectively, with minimal overfitting, indicating strong generalization. The fine-tuned ViT model efficiently captured the temporal and spatial features of PSL gestures, highlighting its ability to classify complex sign language patterns. The following Figures 4 and 5 illustrate the training and validation loss of over ten epochs.

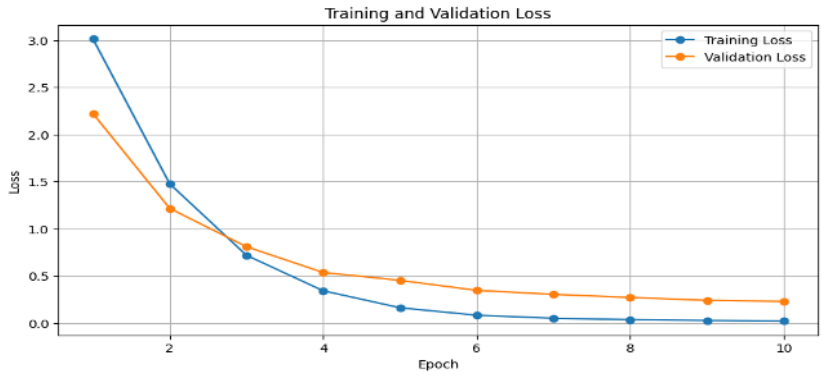

**Figure 4:** Training and Validation Loss Curve Over Epochs.

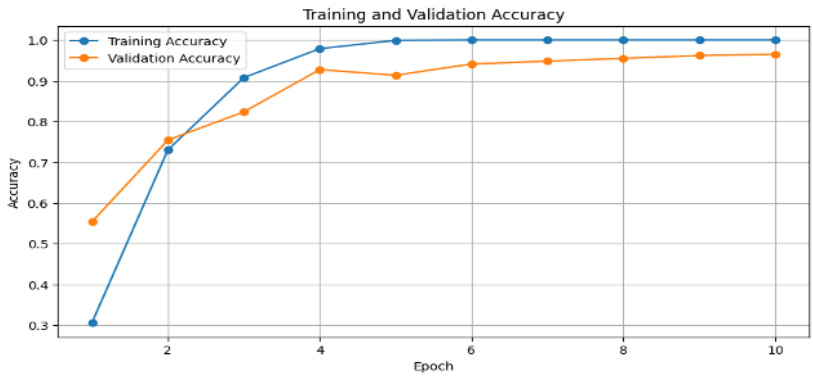

**Figure 5:** Training and Validation Accuracy Curve Over Epochs.

The results indicate that the model can reliably classify PSL gestures with minimal errors. The high f1_score, a balance of precision and recall, highlights the model's robustness and ability to generalize across different gesture classes. In the below Tabel 2 shows the result of performance matrices:

**Table 2.** Result of the Performance Metrics.

| Metric | Value % |
| --- | --- |
| Precision | 97.59% |
| Recall | 97.59% |
| F1_score | 97.58% |

**Table 2:** performance metrics of the ViT-base- patch16-224 model.

Table 3 presents a qualitative comparison with selected prior studies, focusing on target language type rather than dataset size or input modality. This aligns with the scope of our work, which aims to introduce the first domain-specific dataset for mathematical gestures in Palestinian sign language PSL. Unlike previous works that emphasize alphabetic or general gestures in well-supported sign languages such as ASL, ISL, and ArSL, our study contributes novel educational resources addressing a critical gap in the sign language dataset for low-resource communities.

Moreover, while prior studies, such as [21] and [22], concentrated on static alphabet signs, and [23] worked on Isolated words and phrases, our study takes a distinct approach by focusing on mathematical gestures, covering numbers, arithmetic operations, and geometric shapes. This emphasis on mathematical education makes our dataset particularly valuable for enhancing STEM learning accessibility among hard-of-hearing students.

Additionally, our dataset consists of 41 gesture classes, ensuring a diverse and comprehensive representation of mathematical concepts, which is rarely explored in sign language research. This specialization allows for a tailored application in educational settings, making our approach highly practical and impactful for hard-of-hearing students.

**Table 3.** Compassion of sign language datasets with our study.

| Study | Dataset name | Dataset type | classes | Language |
|---|---|---|---|---|
| Our study | Custom dataset | Mathematical gestures | Numbers, operations, and geometric. | Palestinian sign language PSL |
| [23] | WLASL2000 | Isolated words | Words and phrases | American Sign Language ASL |
| [21] | Customs ISL dataset | Static signs letters | Alphabet hand gestures | Indian sign language ISL |
| [22] | ArSl2018 | Static signs letters | Arabic alphabet | Arabic sign language ArSl |

**Table 3:** Compassion of sign language datasets with our study.

## 4 Conclusion and Future Work

This study introduces a domain-specific dataset for Palestinian sign language focused on mathematical gestures, to enhance accessibility for hard-of-hearing students. By training a vision transformer model in 41 gesture classes, we achieved high classification accuracy. Confirming the potential of deep learning in sign language recognition.

Building on this foundation, Future developments will aim to enrich the dataset gesture diversity and participant variation, refine model evaluation using detailed performance metrics (e.g., confusion matrices and per-class accuracy), and explore practical deployment in a real-world educational context. Additionally, expanding the system's recognition capabilities to handle more complex sign context sign structures may further support the creation of interactive and inclusive learning technologies.

## Acknowledgment

We extend our sincere gratitude to Mr. Khalil Al-Alawneh for his invaluable contribution to this research, particularly in the field of Palestinian sign language PSL.